# Text-Based Approaches to Item Alignment to Content Standards in Large-Scale Reading & Writing Tests

Yanbin Fu, Hong Jiao, Tianyi Zhou,
Nan Zhang, Ming Li, Qingshu Xu, Sydney Peters, Robert W. Lissitz

University of Maryland, College Park

**Abstract**

Aligning test items to content standards is a critical step in test development to collect validity evidence based on content. Item alignment has typically been conducted by human experts. This judgmental process can be subjective and time-consuming. This study investigated the performance of fine-tuned small language models (SLMs) for automated item alignment using data from a large-scale standardized reading and writing test for college admissions. Different SLMs were trained for alignment at both domain and skill levels respectively with 10 skills mapped to 4 content domains. The model performance was evaluated in multiple criteria on two testing datasets. The impact of types and sizes of the input data for training was investigated. Results showed that including more item text data led to substantially better model performance, surpassing the improvements induced by sample size increase alone. For comparison, supervised machine learning models were trained using the embeddings from the multilingual-E5-large-instruct model. The study results showed that fine-tuned SLMs consistently outperformed the embedding-based supervised machine learning models, particularly for the more fine-grained skill alignment. To better understand model misclassifications, multiple semantic similarity analysis including pairwise cosine similarity, Kullback-Leibler divergence of embedding distributions, and two-dimension projections of item embeddings were conducted. These analyses consistently showed that certain skills in Test A and Test B were semantically too close, providing evidence for the observed misclassification.

**Keywords:** Automated Item Alignment, Large-Scale Tests, Content Validity, Language Models, BERT

# 1. Introduction

Item alignment is part of alignment defined as the consistency among assessments, content standards, and instructional practices (Smith & O'Day, 1990; Webb, 1997). The degree of item alignment to content standards is critical evidence for validity based on content. If items are not aligned with content standards and expected cognitive complexity, an assessment may fail to represent the construct intended to be measured by the content standards. This ultimately undermines the content validity and leads to invalid inference of test scores. Therefore, item alignment is required to assure the psychometric quality of high-stakes large-scale assessment programs, especially for state assessment programs (Bhola et al., 2003; Embretson & Reise, 2000; Herman et al., 2003; Kane, 2006; Martone & Sireci, 2009; Webb, 1997).

The process of aligning test items with content standards has been referred to different terminologies across different fields. In educational measurement, especially state tests, the term *item alignment* is commonly used to describe the process examining the degree to which an item is mapped to the domains, subdomains, skills, subskills, and standards specified in state *content standards* (Nemeth et al., 2016; Ramesh et al., 2016). In learning analytics, the content standards are referred to as *knowledge component* (KC; Ozyurt et al., 2025) or *content* (Lima et al., 2018), and the task of aligning items to standards are called *KC annotation* (Ozyurt et al., 2025), *KC tagging* (Moore et al., 2024), *KC labeling* (Shen et al., 2021; Tan & Kim, 2024), *content analysis* (Lima et al., 2018), *concept tagging* (Huang et al., 2023), or *question annotation* (Wang et al., 2023). Although these terms differ, they all refer to the same task of aligning test items to content standards.

Item alignment is typically conducted manually by content experts. The process involves reviewing test items one by one and determining which content standards each item measures. Content experts rely on their subject-matter expertise and professional judgment to assess whether the item content is aligned with one or multiple content standards. However, this approach has some limitations. First, manual alignment is time-consuming and labor-intensive, particularly when aligning items in large-scale standardized assessments (Bier et al., 2019; Ding et al., 2025; Zhou & Ostrow, 2022). An alignment workshop typically lasts for 2-3 days for one test form. This process typically involves 6 to 8 panelists, but may include as few as 3 or as many as 30 subject-matter experts, depending on the context (Zhang et al., 2025). Second, expert judgment introduces subjectivity, as panelists may interpret content standards differently. This individual variability can lead to inconsistencies and reduce the reliability of alignment outcomes (Camilli, 2024; Khan et al., 2021). Third, as some test items are developed to assess more complex domains and skills, often involving multiple skills, domains, or hierarchical label structures, manual alignment methods become increasingly challenging (Li et al., 2024).

To address the limitations of manual item alignment, researchers started exploring automated alignment methods using machine learning and natural language processing (NLP) techniques. These approaches aim to enhance consistency, reduce labor, and support scalability in large-scale assessment (see Qu et al., 2011, for an early review). Broadly, automated item alignment methods can be classified into two categories: feature-based approaches and language model-based approaches.

Feature-based approaches can be further divided into two categories: linguistic feature-based models and embedding-based models. Early research used hand-crafted linguistic features such as word counts, keyword overlaps, or term frequency-inverse document frequency (TF-IDF; Sparck Jones, 1972) to capture surface-level textual features. These features were then used in supervised or unsupervised classification models such as K-nearest neighbor (KNN; Karlovcec et al., 2012), support vector machines (SVM; Karlovcec et al., 2012; Yilmazel et al., 2007), Latent Dirichlet Allocation (LDA; Anderson et al., 2020), and XGBoost (Tian et al., 2022). For example, Karlovcec et al. (2012) applied SVM and KNN to align ASSISTments math items with 106 content labels using textual features. Pardos and Dabu (2017) used skip-gram and bag-of-words representations to align items with 198 content labels. However, these linguistic features lacked the ability to capture contextual or sequential information, limiting their ability to represent deeper semantic relationships with content standards.

As neural network models became more prevalent, studies began to explore convolutional neural networks (CNNs; Kim, 2014) for capturing local n-gram features and recurrent neural networks (RNNs; Schuster & Paliwal, 1997) for modeling bidirectional dependencies. RNNs, in particular, excel at processing sequences of arbitrary length and are more commonly used for item alignment. Sun et al. (2018) adopted Bi-Long Short-Term Memory (LSTM), a type of RNN model, to automatically align English questions. Results showed that BiLSTM outperformed classic machine learning models like SVM, achieving an F1-score of 0.562 compared to 0.447.

More recent studies have utilized embeddings as features. These embeddings are either static word embeddings extracted from Word2Vec (Mikolov et al., 2013) or GloVe (Pennington et al., 2014), or contextual embeddings derived from models such as BERT (Devlin et al., 2019). For example, Tian et al. (2022) applied the embeddings from Word2Vec with the Keyphrase extraction features as inputs to XGBoost to align high school math practice items to content standards. This model outperformed baseline models including the vector space model (VSM), SVM, neural networks (NN), and LSTM.

With the advancement of transformer-based language models, small language models such as BERT and RoBERTa have been used for text encoding with contextualized embeddings through self-attention mechanisms. Shen et al. (2021) explored the BERT model for math KC classification. Their results showed that BERT outperformed both the baseline classifiers including SVM, XGBoost, Random Forest, Skip-Gram NN, and MLP as well as the BERT base model without fine-tuning across multiple input settings. Khan et al. (2021) developed Catalog for automatically aligning items with the Next Generation Science Standards (NGSS), leveraging BERT and GPT series to compute semantic similarity between item and standards. Tan and Kim (2024) compared multiple models including XGBoost with embeddings from FastText, fine-tuned BERT-base, BERT-large, RoBERTa-large, and GPT-3.5 with zero-shot prompting. Their results showed that fine-tuned RoBERTa-large consistently outperformed other models. Similarly, Ding et al. (2025) trained a RoBERTa-base model with hierarchical attention to capture hierarchical content structure and multi-label smoothing. The trained model consistently outperformed Bi-LSTM, Bi-GRU, and BERT in math item alignment.

Recently, LLMs such as GPT-3.5 and GPT-4 have been explored for automated item alignment (Li et al., 2024; Liu et al., 2025). These methods rely on carefully designed prompts to guide the LLMs to either classify or generate knowledge labels based on the item text. Wang et al. (2023) used GPT-4 to classify items from the American Board of Family Medicine test into hierarchical categories via zero-shot and few-shot prompting strategies. Li et al. (2024) investigated item alignment as a binary classification task by presenting a math problem and a candidate knowledge description to a LLM, which was then prompted to determine whether the two aligned. Their prompting included knowledge definitions, boundary descriptions, and few-shot examples. Further, a self-reflection mechanism was included to allow the model to re-evaluate and revise its initial prediction. Their experiments using primary school math problems demonstrated that GPT-4 achieved over 90% accuracy, outperforming other LLMs such as GPT-3.5-Turbo, LLAMA2-70B-Chat, Mixtral-8x7B, Qwen1.5-72B, InternLM2-20B-Chat, and InternLM2-20B-Math. Instead of aligning items with predefined content standards, Moore et al. (2024) used GPT-4 to directly generate KC labels, showing that LLMs can simulate expert reasoning and even construct hierarchical knowledge ontologies.

In general, feature-based models rely on either linguistic features (e.g., bag of words and TF-IDF) or embeddings from Word2Vec, Glove, or language models such as BERT or GPT. The embeddings can capture richer semantic information than traditional linguistic features. However, these embeddings are used directly in the downstream classification tasks without further adapted to the tasks. LLMs offer more scalability and eliminate the need for fine-tuning. However, the use of LLMs often raises concerns about data privacy. Most studies on fine-tuning of SLMs were conducted between 2021 and 2022, with limited focus on large-scale state assessment programs. Therefore, a more comprehensive investigation is warranted. Additionally, since SLMs maintain high technical quality while posing minimal data security risks, they merit further extensive research.

To fill critical gaps in the research literature regarding automated item alignment within large-scale state assessment programs, this study examines how SLMs can be optimally fine-tuned for item alignment applications in state assessment contexts. The research questions guiding this study are outlined below.

1. To what extent do training sample size and input representation (e.g., question stem, options, rationale) influence the alignment accuracy of fine-tuned SLMs?
2. How do different SLMs perform in aligning test items to skill and domain categories?
3. Which skills are most prone to misalignment, and what semantic characteristics contribute to misclassification errors?

## 2. Methods

### 2.1 Data

This study utilized 1,270 items from a large-scale reading and writing test (labeled as Test A). 80% of the total items were used to train the models while 20% were held out as a test dataset. In addition, 1,052 items from another large-scale reading and writing test (labeled as Test B) were used as an external test dataset to assess model generalizability as Test B covers the same content domains and standards. Each item consists of a prompt or lead-in such as a reading passage or a context, a question or stem with four answer options, a key or the correct answer, and a rationale that explains the correct option and incorrect options. Some items may also contain a graph or table. The graphs were described verbally. The description was extracted from the HTML alt text, provided for accessibility accommodation. Tables in items were converted to LaTeX via GPT-4. Each item contains a domain label and a skill label. The hierarchy between domains and skills is presented in Table 1. Brief descriptions of the four domains are provided in Appendix Table A.1. Table 1 presents the domains and skills across Test A and Test B item sets. There are four domains and ten skills in each of the tests. The four domains are *Standard English Conventions*, *Information and Ideas*, *Expression of Ideas*, and *Craft and Structure*. The ten skills include *Boundaries*, *Form, Structure, and Sense*, *Command of Evidence*, *Inferences*, *Central Ideas and Details*, *Transitions*, *Rhetorical Synthesis*, *Words in Context, Text Structure and Purpose*, and *Cross-Text Connections*. The number of items in each domain and skill varies between Test A and Test B datasets. While the distribution of items across domains is relatively balanced, the distribution across skills is less balanced, with the fewest items in *Cross-Text Connections*.

**Table 1**

*Content Domains and Skills and their Respective Sample Sizes*

| Domains | Test A | Test B | Skills | Test A | Test B |
|---|---|---|---|---|---|
| Standard English Conventions | 289 | 253 | Boundaries | 148 | 127 |
| | | | Form, Structure, and Sense | 141 | 126 |
| Information and Ideas | 383 | 307 | Command of Evidence | 196 | 160 |
| | | | Inferences | 93 | 70 |
| | | | Central Ideas and Details | 94 | 77 |
| Expression of Ideas | 269 | 228 | Transitions | 128 | 109 |
| | | | Rhetorical Synthesis | 141 | 119 |
| Craft and Structure | 329 | 264 | Words in Context | 180 | 151 |
| | | | Text Structure and Purpose | 98 | 77 |
| | | | Cross-Text Connections | 51 | 36 |
| Total | 1270 | 1052 | 10 | 1270 | 1052 |

**2.2 Impact of Sample Sizes and Input Data on Model Training**

To investigate the impact of sample size and input data on item alignment accuracy, the study experimented with different sample sizes and input data in the training dataset. BERT-base was first used for such exploration. Specifically, the total sample size was manipulated at 500, 750, 1000, or 1270 items. Nine input data types are listed below:

1. Prompt only
2. Prompt+table+figure
3. Prompt+table+figure+options
4. Prompt+table+figure+options+key
5. Prompt+table+figure+options+key+rationale
6. Prompt+table+figure+question text
7. Prompt+table+figure+question text+options
8. Prompt+table+figure+question text+options+key
9. Prompt+table+figure+question text+options+key+rationale

**2.3 Models**

To evaluate SLMs performance in item alignment, SLMs were fine-tuned. This study explored both SLM-based modeling approaches and embedding feature-based supervised machine learning models. The 12 fine-tuned SLMs include BERT-base, BERT-large (Devlin et al., 2019), ALBERT-base (Lan et al., 2019), DistilBERT-base (Sanh et al., 2019), All-DistilRoBERTa (Liu et al., 2019; Sanh et al., 2019), ELECTRA-small, ELECTRA-base (Clark et al., 2020), RoBERTa-base, RoBERTa-large (Liu et al., 2019), DeBERTa-base (He et al., 2020), DeBERTa-large (He et al., 2021), and ConvBERT (Jiang et al., 2020). All models were fine-tuned using standard classification heads and optimized using cross-entropy loss.

For comparison, embedding-based approaches were explored as well. The multilingual-E5-large-instruct model (Wang et al., 2024) was used to encode each item into a fixed-length sentence embedding. This model achieved strong performance on multiple embedding tasks in the Massive Text Embedding Benchmark (MTEB; Muennighoff et al., 2022). These embeddings were derived for both the training and testing sets using the model's [CLS] token output. Multiple supervised machine learning models, including logistic regression, SVM, Naive Bayes, Random Forest, Gradient Boosting, XGBoost, LightGBM, MLP, and KNN, were trained for aligning items to domains and skills, respectively.

**2.4 Model Fine-Tuning**

The Test A data was split into 60% training, 20% validation, and 20% testing. The models were fine-tuned on the Test A training dataset. Checkpoints were selected based on the highest validation accuracy. Model performance was evaluated on the Test A test set and additionally tested on the Test B dataset to assess model generalizability. In addition, models were trained with 15 epochs using the AdamW optimizer, a learning rate of 2e-5, a batch size of 8, and a linear learning rate scheduler with a warmup ratio of 0.1. Each SLM was fine-tuned separately for domain and skill alignment. Item input texts were tokenized using the tokenizer of each SLM and truncated to a maximum length of 512 tokens. It is worth noting that when the rationale was included as part of the input, approximately 20% of items had truncated input data. This indicates that, in those cases, the rationale was not fully utilized as part of input for model training.

The model performance was evaluated in terms of accuracy, recall, precision, weighted F1 score, and Cohen's kappa coefficient on both the Test A's test set and the Test B items. These metrics assess different aspects of prediction accuracy.

**2.5 Embedding-Based Analysis of Misclassification**

To investigate patterns of model misclassification for skill alignment, this study employed a range of embedding-based analytical methods. These methods included exploration of semantic similarity, representational overlap, and distributional structure for explaining skill alignment errors. Embedding-level similarity measures including cosine similarity and Kullback-Leibler (KL) divergence were computed. Visualization was presented to highlight overlapping semantic representations that may contribute to these classification errors.

This study calculated all-pairwise cosine similarity between the selected skill groups with high rates of misclassification to examine their semantic proximity in the embedding space. Given two sets of embeddings, $A = \{a_1, \ldots, a_m\}$ and $B = \{b_1, \ldots, b_n\}$, where m and n represent the number of items in the two skill sets respectively, the average cosine similarity is defined in equation 6. Higher scores indicate greater similarity between two item embedding, potentially reflecting a source of confusion for the model.

$$AvgCos(A, B) = \frac{1}{mn} \sum_{i=1}^{m} \sum_{j=1}^{n} cos(\theta_{ij}) = \frac{1}{mn} \sum_{i=1}^{m} \sum_{j=1}^{n} \frac{a_i \cdot b_j}{\|a_i\| \cdot \|b_j\|}. \tag{1}$$

To visualize the structure of the embeddings, this study applied three common dimensionality reduction techniques, including principal component analysis (PCA), t-distributed stochastic neighbor embedding (t-SNE), and isometric mapping (ISOMAP), to project item-level embeddings from selected models into a two-dimension space with points color-coded by their skill or domain labels.

In addition, this study computed the KL divergence between the embedding distributions to interpret the misclassification. Specifically, embeddings extracted by the multilingual-E5-large-instruct were projected into a two-dimensional space using PCA, and skill-specific distributions were estimated using two-dimensional histograms with Gaussian kernel smoothing. For each skill pair, the KL divergence from distribution Q to distribution P is defined as in equation 7.

$$D_{KL}(P\|Q) = \sum_i P(i)log(\frac{P(i)}{Q(i)}). \quad (2)$$

where *P(i)* and *Q(i)* denote the normalized density estimates in the $i^{th}$ bin of the 2-dimensional histograms for two categories of items. Lower KL divergence values indicate greater similarity between the embedding distributions of two items measuring different skills, which may help explain misclassifications between certain skill pairs.

## 3. Results
### 3.1 Impact of Sample Sizes and Input Data on Model Training

The effects of sample size and input data on the performance of the BERT-base model in both skill and domain alignment tasks are summarized in Tables 2 and 3. Across all sample sizes, the richness of the input data had a substantial impact on alignment performance. Specifically, "prompt_only" containing only minimal information yielded the lowest scores across all evaluation metrics. As additional input components were added, such as tables, figures, options, keys, rationales, and question text, model performance improved. For example, in skill alignment using 400 out of 500 items for model training, the weighted F1 score increased from 0.664 under the "prompt_only" condition to 0.919 under the condition using all input sources except item question.

Notably, the inclusion of the question text resulted in a sharp jump in performance. For example, in skill alignment with 600 training items, the model's weighted F1 increased from 0.754 with "prompt_ table_figure" to 0.947 with "prompt_table_figure_qtext". Based on this initial exploration, it was found that the question texts for some items within the same domain or skill category often share highly repetitive question templates. For example, all items under the domain category *Standard English Conventions*, contained identical question text as "*Which choice completes the text so that it conforms to the conventions of Standard English?*". More example questions for different domains and skills can be found in Appendix Table A.2. Such question templates may allow a model to achieve high prediction performance by memorizing superficial text patterns, rather than truly learning the semantic relationship between the item content and its corresponding domain or skill. The inclusion of item questions may introduce systematic patterns for some particular domains or skills. This form of shortcut learning may lead to overfitting, preventing the model from fully leveraging the textual information present in other components of the item. To mitigate the shortcut issue, the question text was removed from the training dataset. Therefore, this study included only the following textual components from each item: the prompt, the four options, key, and the rationale. If an item contained a table or a figure, they were included as well. These components were concatenated to form the input data used for model training and comparison. An example input data is provided in Appendix Table A.3.

In contrast, increasing the training sample size from 400 to 1016 yielded modest gains in performance, particularly when compared to the gains achieved through adding more input components. For example, in skill alignment with "prompt_only", the weighted F1 score improved from 0.664 with 400 items to 0.798 with 800 items, whereas the same level of performance improvement could be surpassed by adding table/figure with 400 items to a F1 score of 0.801. Domain alignment followed a similar pattern, though with generally higher performance. Under the "prompt_only" condition, a weighted F1 score of 0.919 was achieved with just 400 items for domain alignment. Even higher performance (0.927) was observed when all input components were included, even with only 750 items. These findings suggest that while larger training samples improve model performance, incorporating richer contextual information in the input data has an even greater impact. Overall, the evidence highlights the importance of more input data from each item in achieving high-quality alignment, especially when the number of items for training is limited.

**Table 2**

*Performance of BERT-base across Training Sample Sizes and Input Data for Skill Alignment*

| Sample Sizes | Input Conditions | Accuracy | Precision | Recall | Weighted F1 | Cohen's Kappa |
|---|---|---|---|---|---|---|
| 400 | prompt_only | 0.700 | 0.690 | 0.700 | 0.664 | 0.662 |
| | prompt_table_figure | 0.810 | 0.813 | 0.810 | 0.801 | 0.786 |
| | prompt_table_figure_options | 0.900 | 0.904 | 0.900 | 0.897 | 0.886 |
| | prompt_table_figure_options_key | 0.880 | 0.886 | 0.880 | 0.876 | 0.864 |
| | prompt_table_figure_options_key_rationale | 0.920 | 0.926 | 0.920 | 0.919 | 0.909 |
| | prompt_table_figure_qtext | 0.890 | 0.915 | 0.890 | 0.893 | 0.876 |
| | prompt_table_figure_qtext_options | 1.000 | 1.000 | 1.000 | 1.000 | 1.000 |
| | prompt_table_figure_qtext_options_key | 0.980 | 0.984 | 0.980 | 0.981 | 0.977 |
| | prompt_table_figure_qtext_options_key_rationale | 0.940 | 0.970 | 0.940 | 0.935 | 0.932 |
| 600 | prompt_only | 0.787 | 0.796 | 0.787 | 0.787 | 0.760 |
| | prompt_table_figure | 0.767 | 0.795 | 0.767 | 0.754 | 0.738 |
| | prompt_table_figure_options | 0.880 | 0.876 | 0.880 | 0.871 | 0.865 |
| | prompt_table_figure_options_key | 0.900 | 0.911 | 0.900 | 0.898 | 0.887 |
| | prompt_table_figure_options_key_rationale | 0.933 | 0.948 | 0.933 | 0.932 | 0.925 |
| | prompt_table_figure_qtext | 0.947 | 0.948 | 0.947 | 0.947 | 0.940 |
| | prompt_table_figure_qtext_options | 0.993 | 0.994 | 0.993 | 0.993 | 0.992 |
| | prompt_table_figure_qtext_options_key | 0.980 | 0.980 | 0.980 | 0.980 | 0.977 |

| Sample Sizes | Input Conditions | Accuracy | Precision | Recall | Weighted F1 | Cohen's Kappa |
|---|---|---|---|---|---|---|
| | prompt_table_figure_qtext_options_key_rationale | 0.980 | 0.982 | 0.980 | 0.980 | 0.977 |
| 800 | prompt_only | 0.800 | 0.817 | 0.800 | 0.798 | 0.777 |
| | prompt_table_figure | 0.815 | 0.812 | 0.815 | 0.811 | 0.793 |
| | prompt_table_figure_options | 0.865 | 0.887 | 0.865 | 0.871 | 0.849 |
| | prompt_table_figure_options_key | 0.890 | 0.915 | 0.890 | 0.896 | 0.877 |
| | prompt_table_figure_options_key_rationale | 0.850 | 0.883 | 0.850 | 0.855 | 0.832 |
| | prompt_table_figure_qtext | 0.950 | 0.950 | 0.950 | 0.950 | 0.944 |
| | prompt_table_figure_qtext_options | 0.990 | 0.990 | 0.990 | 0.990 | 0.989 |
| | prompt_table_figure_qtext_options_key | 0.995 | 0.995 | 0.995 | 0.995 | 0.994 |
| | prompt_table_figure_qtext_options_key_rationale | 0.995 | 0.995 | 0.995 | 0.995 | 0.994 |
| 1016 | prompt_only | 0.795 | 0.794 | 0.795 | 0.789 | 0.770 |
| | prompt_table_figure | 0.803 | 0.803 | 0.803 | 0.784 | 0.778 |
| | prompt_table_figure_options | 0.890 | 0.910 | 0.890 | 0.890 | 0.876 |
| | prompt_table_figure_options_key | 0.902 | 0.904 | 0.902 | 0.900 | 0.889 |
| | prompt_table_figure_options_key_rationale | 0.909 | 0.913 | 0.909 | 0.909 | 0.898 |
| | prompt_table_figure_qtext | 0.945 | 0.945 | 0.945 | 0.945 | 0.938 |
| | prompt_table_figure_qtext_options | 0.984 | 0.985 | 0.984 | 0.984 | 0.982 |
| | prompt_table_figure_qtext_options_key | 0.992 | 0.993 | 0.992 | 0.992 | 0.991 |
| | prompt_table_figure_qtext_options_key_rationale | 1.000 | 1.000 | 1.000 | 1.000 | 1.000 |

**Table 3**

*Performance of BERT-base across Training Sample Sizes and Input Data for Domain Alignment*

| Sample Sizes | Input Conditions | Accuracy | Precision | Recall | Weighted F1 | Cohen's Kappa |
|---|---|---|---|---|---|---|
| 400 | prompt_only | 0.920 | 0.929 | 0.920 | 0.919 | 0.891 |
| | prompt_table_figure | 0.930 | 0.931 | 0.930 | 0.930 | 0.905 |
| | prompt_table_figure_options | 0.960 | 0.963 | 0.960 | 0.960 | 0.945 |
| | prompt_table_figure_options_key | 0.970 | 0.973 | 0.970 | 0.970 | 0.959 |
| | prompt_table_figure_options_key_rationale | 0.990 | 0.990 | 0.990 | 0.990 | 0.986 |
| | prompt_table_figure_qtext | 1.000 | 1.000 | 1.000 | 1.000 | 1.000 |
| | prompt_table_figure_qtext_options | 0.970 | 0.973 | 0.970 | 0.970 | 0.959 |
| | prompt_table_figure_qtext_options_key | 1.000 | 1.000 | 1.000 | 1.000 | 1.000 |
| | prompt_table_figure_qtext_options_key_rationale | 0.980 | 0.981 | 0.980 | 0.980 | 0.973 |
| 600 | prompt_only | 0.900 | 0.900 | 0.900 | 0.900 | 0.866 |
| | prompt_table_figure | 0.900 | 0.902 | 0.900 | 0.899 | 0.866 |
| | prompt_table_figure_options | 0.953 | 0.958 | 0.953 | 0.954 | 0.937 |
| | prompt_table_figure_options_key | 0.953 | 0.960 | 0.953 | 0.954 | 0.937 |
| | prompt_table_figure_options_key_rationale | 0.927 | 0.934 | 0.927 | 0.927 | 0.902 |
| | prompt_table_figure_qtext | 1.000 | 1.000 | 1.000 | 1.000 | 1.000 |
| | prompt_table_figure_qtext_options | 1.000 | 1.000 | 1.000 | 1.000 | 1.000 |
| | prompt_table_figure_qtext_options_key | 1.000 | 1.000 | 1.000 | 1.000 | 1.000 |
| | prompt_table_figure_qtext_options_key_rationale | 0.987 | 0.987 | 0.987 | 0.987 | 0.982 |
| 800 | prompt_only | 0.885 | 0.888 | 0.885 | 0.885 | 0.846 |
| | prompt_table_figure | 0.900 | 0.901 | 0.900 | 0.900 | 0.866 |
| | prompt_table_figure_options | 0.965 | 0.966 | 0.965 | 0.965 | 0.953 |
| | prompt_table_figure_options_key | 0.960 | 0.962 | 0.960 | 0.960 | 0.947 |
| | prompt_table_figure_options_key_rationale | 0.940 | 0.947 | 0.940 | 0.941 | 0.920 |
| | prompt_table_figure_qtext | 1.000 | 1.000 | 1.000 | 1.000 | 1.000 |
| | prompt_table_figure_qtext_options | 1.000 | 1.000 | 1.000 | 1.000 | 1.000 |
| | prompt_table_figure_qtext_options_key | 1.000 | 1.000 | 1.000 | 1.000 | 1.000 |
| | prompt_table_figure_qtext_options_key_rationale | 0.990 | 0.990 | 0.990 | 0.990 | 0.987 |
| 1016 | prompt_only | 0.898 | 0.901 | 0.898 | 0.897 | 0.860 |
| | prompt_table_figure | 0.898 | 0.908 | 0.898 | 0.896 | 0.860 |
| | prompt_table_figure_options | 0.965 | 0.967 | 0.965 | 0.965 | 0.952 |
| | prompt_table_figure_options_key | 0.976 | 0.977 | 0.976 | 0.976 | 0.968 |

| | | | | | |
|---|---|---|---|---|---|
| prompt_table_figure_options_key_rationale | 0.980 | 0.981 | 0.980 | 0.980 | 0.973 |
| prompt_table_figure_qtext | 0.996 | 0.996 | 0.996 | 0.996 | 0.995 |
| prompt_table_figure_qtext_options | 0.992 | 0.992 | 0.992 | 0.992 | 0.989 |
| prompt_table_figure_qtext_options_key | 1.000 | 1.000 | 1.000 | 1.000 | 1.000 |
| prompt_table_figure_qtext_options_key_rationale | 1.000 | 1.000 | 1.000 | 1.000 | 1.000 |

### 3.2 Model Performance Comparison

Tables 6 and 7 summarize the model performance on Test A skill and domain alignment for both fine-tuned SLMs and embedding-based supervised machine learning models. In general, the fine-tuned SLMs yielded much higher alignment accuracy than the embedding-based models across all metrics. In skill alignment, fine-tuned ConvBERT and RoBERTa-large achieved perfect performance (all metrics = 1.000). Even the weakest ALBERT-base still performed well (precision = 0.949, recall = 0.945, accuracy = 0.945, weighted F1 = 0.943, and Cohen's kappa = 0.938). In contrast, embedding-based models achieved F1 scores ranging from 0.513 to 0.829. Among them, MLP performed best.

For domain alignment, fine-tuned models still consistently outperformed the embedding-based classifiers. ConvBERT, RoBERTa-large, RoBERTa-base, and DeBERTa-base again achieved perfect performance (all metrics = 1.000), while BERT-base, BERT-large, and ELECTRA-base performed slightly worse, still with all evaluation metrics above 0.995. All embedding-based models also performed significantly better on domain alignment than on skill alignment, with MLP achieving a weighted F1 score of 0.921. This suggests that domain alignment is a comparatively easier alignment task.

**Table 6**
*Model Performance on* Test *A Skill Alignment.*

| Model | Precision | Recall | Accuracy | Weighted F1 | Cohen's Kappa |
|---|---|---|---|---|---|
| BERT-base | 0.996 | 0.996 | 0.996 | 0.996 | 0.996 |
| BERT-large | 0.989 | 0.988 | 0.988 | 0.988 | 0.987 |
| ALBERT-base | 0.949 | 0.945 | 0.945 | 0.943 | 0.938 |
| ConvBERT | 1.000 | 1.000 | 1.000 | 1.000 | 1.000 |
| All-DistilRoBERTa | 0.985 | 0.984 | 0.984 | 0.984 | 0.982 |
| ELECTRA-base | 0.992 | 0.992 | 0.992 | 0.992 | 0.991 |
| ELECTRA-small | 0.974 | 0.969 | 0.969 | 0.966 | 0.965 |
| RoBERTa-base | 0.996 | 0.996 | 0.996 | 0.996 | 0.996 |
| RoBERTa-large | 1.000 | 1.000 | 1.000 | 1.000 | 1.000 |
| DeBERTa-base | 0.985 | 0.984 | 0.984 | 0.984 | 0.982 |
| DeBERTa-large | 0.996 | 0.996 | 0.996 | 0.996 | 0.996 |
| DistilBERT-base | 0.992 | 0.992 | 0.992 | 0.992 | 0.991 |
| Logistic Regression | 0.538 | 0.646 | 0.646 | 0.563 | 0.593 |
| SVM | 0.642 | 0.701 | 0.701 | 0.643 | 0.658 |
| Naive Bayes | 0.764 | 0.744 | 0.744 | 0.749 | 0.713 |
| Random Forest | 0.591 | 0.610 | 0.571 | 0.513 | 0.554 |
| Gradient Boosting | 0.575 | 0.583 | 0.594 | 0.573 | 0.526 |
| XGBoost | 0.618 | 0.610 | 0.610 | 0.597 | 0.560 |
| LightGBM | 0.652 | 0.665 | 0.665 | 0.643 | 0.621 |
| MLP | 0.816 | 0.823 | 0.835 | 0.829 | 0.800 |
| KNN | 0.524 | 0.535 | 0.535 | 0.513 | 0.476 |

**Table 7**
*Model Performance on Test A Domain Alignment.*

| Model | Precision | Recall | Accuracy | Weighted F1 | Cohen's Kappa |
|---|---|---|---|---|---|
| BERT-base | 0.996 | 0.996 | 0.996 | 0.996 | 0.995 |
| BERT-large | 0.996 | 0.996 | 0.996 | 0.996 | 0.995 |
| ALBERT-base | 0.967 | 0.965 | 0.965 | 0.965 | 0.952 |
| ConvBERT | 1.000 | 1.000 | 1.000 | 1.000 | 1.000 |
| All-DistilRoBERTa | 0.996 | 0.996 | 0.965 | 0.965 | 0.995 |
| ELECTRA-base | 0.996 | 0.996 | 0.996 | 0.996 | 0.995 |
| ELECTRA-small | 0.980 | 0.980 | 0.980 | 0.980 | 0.973 |
| RoBERTa-base | 1.000 | 1.000 | 1.000 | 1.000 | 1.000 |
| RoBERTa-large | 1.000 | 1.000 | 1.000 | 1.000 | 1.000 |
| DeBERTa-base | 1.000 | 1.000 | 1.000 | 1.000 | 1.000 |
| DeBERTa-large | 0.996 | 0.996 | 0.996 | 0.996 | 0.995 |
| DistilBERT-base | 0.992 | 0.992 | 0.992 | 0.992 | 0.989 |
| Logistic Regression | 0.879 | 0.878 | 0.878 | 0.878 | 0.834 |
| SVM | 0.901 | 0.894 | 0.894 | 0.894 | 0.857 |
| Naive Bayes | 0.839 | 0.827 | 0.827 | 0.827 | 0.767 |
| Random Forest | 0.812 | 0.807 | 0.783 | 0.781 | 0.735 |
| Gradient Boosting | 0.852 | 0.850 | 0.846 | 0.846 | 0.796 |
| XGBoost | 0.829 | 0.823 | 0.823 | 0.824 | 0.760 |
| LightGBM | 0.848 | 0.846 | 0.846 | 0.847 | 0.792 |
| MLP | 0.923 | 0.921 | 0.921 | 0.921 | 0.893 |
| KNN | 0.727 | 0.724 | 0.724 | 0.719 | 0.627 |

Tables 8 and 9 present the SLMs performance on the Test B skill and domain alignment compared with the embedding-based supervised machine learning models. For skill alignment, fine-tuned SLMs still outperformed embedding-based classic machine learning models, although several fine-tuned models exhibited performance drops compared to their performance on the Test A items. ELECTRA-base and RoBERTa-large continued to perform the best (all metrics ≥ 0.993), while DeBERTa-base and ALBERT-base also maintained good performance (all metrics ≥ 0.956).

For Test B domain alignment, DeBERTa-base achieved the highest performance (all metrics = 0.997), while RoBERTa-base and RoBERTa-large also performed strongly (all metrics = 0.994). Other fine-tuned models, including BERT-large and All-DistilRoBERTa, maintained solid performance, with all metrics exceeding 0.98.

These findings indicate that fine-tuned SLMs on Test A items were able to generalize across similar assessment items from Test B. However, the observed drops in performance for certain skills suggest that items in some content categories may be too similar to be aligned properly to its own category. To further investigate this issue, we conducted a misclassification analysis to examine where and why the models failed, particularly focusing on skill-level misclassification.

**Table 8**
*Model Performance on* Test *B Skill Alignment.*

| Model | Precision | Recall | Accuracy | Weighted F1 | Cohen's Kappa |
|---|---|---|---|---|---|
| BERT-base | 0.935 | 0.894 | 0.894 | 0.878 | 0.879 |
| BERT-large | 0.906 | 0.827 | 0.827 | 0.797 | 0.802 |
| ALBERT-base | 0.969 | 0.961 | 0.961 | 0.961 | 0.956 |
| ConvBERT | 0.902 | 0.887 | 0.887 | 0.870 | 0.871 |
| All-DistilRoBERTa | 0.931 | 0.907 | 0.907 | 0.887 | 0.895 |
| ELECTRA-base | 0.993 | 0.993 | 0.993 | 0.993 | 0.993 |
| ELECTRA-small | 0.744 | 0.760 | 0.760 | 0.722 | 0.728 |
| RoBERTa-base | 0.959 | 0.942 | 0.942 | 0.929 | 0.935 |
| RoBERTa-large | 0.994 | 0.994 | 0.994 | 0.994 | 0.994 |
| DeBERTa-base | 0.978 | 0.976 | 0.976 | 0.976 | 0.973 |
| DeBERTa-large | 0.927 | 0.894 | 0.894 | 0.868 | 0.879 |
| DistilBERT-base | 0.940 | 0.920 | 0.920 | 0.910 | 0.910 |
| Logistic Regression | 0.708 | 0.723 | 0.723 | 0.653 | 0.682 |
| SVM | 0.861 | 0.804 | 0.804 | 0.763 | 0.776 |
| Naive Bayes | 0.862 | 0.853 | 0.853 | 0.855 | 0.834 |
| Random Forest | 0.938 | 0.933 | 0.920 | 0.919 | 0.924 |
| Gradient Boosting | 0.881 | 0.879 | 0.883 | 0.882 | 0.864 |
| XGBoost | 0.917 | 0.914 | 0.914 | 0.914 | 0.903 |
| LightGBM | 0.938 | 0.937 | 0.937 | 0.936 | 0.929 |
| MLP | 0.963 | 0.963 | 0.961 | 0.961 | 0.958 |
| KNN | 0.695 | 0.695 | 0.695 | 0.687 | 0.655 |

**Table 9**
*Model Performance on* Test *B Domain Alignment.*

| Model | Precision | Recall | Accuracy | Weighted F1 | Cohen's Kappa |
|---|---|---|---|---|---|
| BERT-base | 0.947 | 0.934 | 0.934 | 0.934 | 0.912 |
| BERT-large | 0.986 | 0.985 | 0.985 | 0.985 | 0.980 |
| ALBERT-base | 0.892 | 0.820 | 0.820 | 0.803 | 0.762 |
| ConvBERT | 0.971 | 0.967 | 0.967 | 0.967 | 0.956 |
| All-DistilRoBERTa | 0.986 | 0.986 | 0.986 | 0.986 | 0.981 |
| ELECTRA-base | 0.928 | 0.904 | 0.904 | 0.902 | 0.872 |
| ELECTRA-small | 0.949 | 0.937 | 0.937 | 0.937 | 0.916 |
| RoBERTa-base | 0.994 | 0.994 | 0.994 | 0.994 | 0.992 |
| RoBERTa-large | 0.994 | 0.994 | 0.994 | 0.994 | 0.992 |
| DeBERTa-base | 0.997 | 0.997 | 0.997 | 0.997 | 0.996 |
| DeBERTa-large | 0.988 | 0.988 | 0.988 | 0.988 | 0.983 |
| DistilBERT-base | 0.940 | 0.926 | 0.926 | 0.925 | 0.901 |
| Logistic Regression | 0.899 | 0.898 | 0.898 | 0.899 | 0.864 |
| SVM | 0.934 | 0.933 | 0.933 | 0.933 | 0.911 |
| Naive Bayes | 0.860 | 0.857 | 0.857 | 0.857 | 0.810 |
| Random Forest | 0.959 | 0.959 | 0.953 | 0.953 | 0.945 |
| Gradient Boosting | 0.959 | 0.959 | 0.958 | 0.958 | 0.945 |
| XGBoost | 0.968 | 0.968 | 0.968 | 0.968 | 0.957 |
| LightGBM | 0.969 | 0.969 | 0.969 | 0.969 | 0.958 |
| MLP | 0.964 | 0.964 | 0.963 | 0.963 | 0.952 |
| KNN | 0.799 | 0.798 | 0.798 | 0.796 | 0.730 |

### 3.3 Exploration of Misclassification

Table 10 presents the weighted F1 scores for each SLM across skill labels in the Test B dataset. Several models, including BERT-base, BERT-large, ConvBERT, All-DistilRoBERTa, ELECTRA-small, RoBERTa-base, DeBERTa-large, and DistilBERT-base exhibited evident drops in F1 scores on some skills. In particular, skill 4 (*Inferences*) and skill 5 (*Central Ideas and Details*) showed consistently low accuracy, with weighted F1 scores below 0.6. As further shown in the confusion matrix analysis (Appendix B), these items were frequently misclassified as Skill 8 (Words in Context), especially by models like BERT-base.

**Table 10**
*Performance of Fine-Tuned Small Language Models on Skill Alignment of Test B Items.*

| Model | Skill 1 | Skill 2 | Skill 3 | Skill 4 | Skill 5 | Skill 6 | Skill 7 | Skill 8 | Skill 9 | Skill 10 |
|---|---|---|---|---|---|---|---|---|---|---|
| BERT-base | 0.996 | 0.992 | 0.997 | 0.692 | **0.250** | 0.991 | 1.000 | 0.737 | 0.924 | 0.986 |
| BERT-large | 0.992 | 0.992 | 0.981 | **0.200** | **0.075** | 1.000 | 0.996 | 0.630 | 0.672 | 1.000 |
| ALBERT-base | 0.988 | 0.976 | 0.968 | 0.824 | 0.797 | 0.995 | 1.000 | 0.993 | 0.981 | 1.000 |
| ConvBERT | 0.992 | 0.996 | 0.972 | **0.150** | 0.678 | 1.000 | 1.000 | 0.741 | 0.900 | 1.000 |
| All-DistilRoBERTa | 0.992 | 0.992 | 0.963 | **0.108** | 0.683 | 0.926 | 1.000 | 0.937 | 0.917 | 0.986 |
| ELECTRA-base | 0.988 | 0.992 | 0.997 | 1.000 | 0.974 | 1.000 | 1.000 | 0.993 | 0.987 | 1.000 |
| ELECTRA-small | 0.988 | 0.988 | 0.672 | **0.000** | **0.000** | 1.000 | 1.000 | 0.619 | 0.653 | 0.839 |
| RoBERTa-base | 0.992 | 0.992 | 0.955 | **0.333** | 0.774 | 1.000 | 1.000 | 0.997 | 0.993 | 1.000 |
| RoBERTa-large | 0.996 | 0.996 | 0.991 | 0.986 | 0.974 | 1.000 | 1.000 | 0.997 | 1.000 | 1.000 |
| DeBERTa-base | 0.996 | 0.996 | 0.984 | 0.867 | 0.900 | 0.995 | 1.000 | 0.984 | 0.980 | 1.000 |
| DeBERTa-large | 0.992 | 0.984 | 1.000 | **0.056** | 0.798 | 0.995 | 1.000 | 0.800 | 0.695 | 1.000 |
| DistilBERT-base | 0.996 | 0.996 | 0.991 | 0.824 | **0.424** | 0.995 | 0.996 | 0.904 | 0.746 | 1.000 |

*Note.* Skill 1 = Boundaries; Skill 2 = Form, Structure, and Sense; Skill 3 = Command of Evidence; Skill 4 = Inferences; Skill 5 = Central Ideas and Details; Skill 6 = Transitions; Skill 7 = Rhetorical Synthesis; Skill 8 = Words in Context; Skill 9 = Text Structure and Purpose; Skill 10 = Cross-Text Connections.

To further investigate misclassifications, this study conducted a series of similarity analyses. Pairwise cosine similarities were calculated between the embeddings of Test A and Test B items labeled as Skill 4, Skill 5, and Skill 8. Table 11 presents the mean pairwise cosine similarities for each pairwise skills. The cosine similarity between Test A Skill 4 and Test A Skill 8 embeddings was 0.827, while the corresponding value for Test B Skill 4 and Skill 8 was 0.828. Similarly, the embeddings for Skill 5 and Skill 8 showed high cosine similarity, with values of 0.825 in Test A and 0.823 in Test B. When comparing across assessments, the cosine similarity between Test A Skill 4 and Test B Skill 4 was 0.824; between Test A Skill 5 and Test B Skill 5 it was 0.822; and between Test A Skill 8 and Test B Skill 5 it was 0.821. These high similarity values indicate that Skills 4, 5, and 8 are semantically close in the embedding space, which may account for the models' difficulty in distinguishing among them.

**Table 11**

*Mean Pairwise Cosine Similarity between Skill 4/5 and Skill 8 across* Test *A and* Test *B Items*

| Pairwise Mapping | | Cosine Similarity |
| --- | --- | --- |
| Between Dataset | | |
| Test A Skill 4 | Test B Skill 4 | 0.824 |
| Test A Skill 4 | Test B Skill 8 | 0.827 |
| Test A Skill 5 | Test B Skill 5 | 0.822 |
| Test A Skill 5 | Test B Skill 8 | 0.826 |
| Test A Skill 8 | Test B Skill 4 | 0.821 |
| Test A Skill 8 | Test B Skill 5 | 0.821 |
| Test A Skill 8 | Test B Skill 8 | 0.838 |
| Within Dataset | | |
| Test A Skill 4 | Test A Skill 8 | 0.827 |
| Test A Skill 5 | Test A Skill 8 | 0.825 |
| Test B Skill 4 | Test B Skill 8 | 0.828 |
| Test B Skill 5 | Test B Skill 8 | 0.823 |

*Note.* Skill 4 = Inferences; Skill 5 = Central Ideas and Details; Skill 8 = Words in Context.

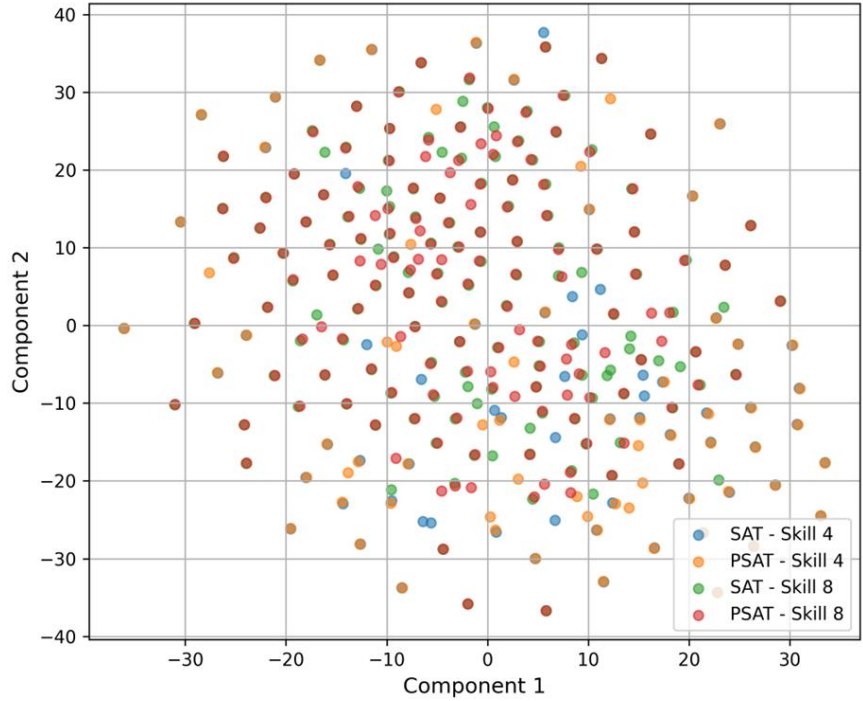

**Figure 1.** *t-SNE Plot of Embeddings for Skill 4 (Inferences) vs. Skill 8 (Words in Context) for* Test *A and* Test *B Items*

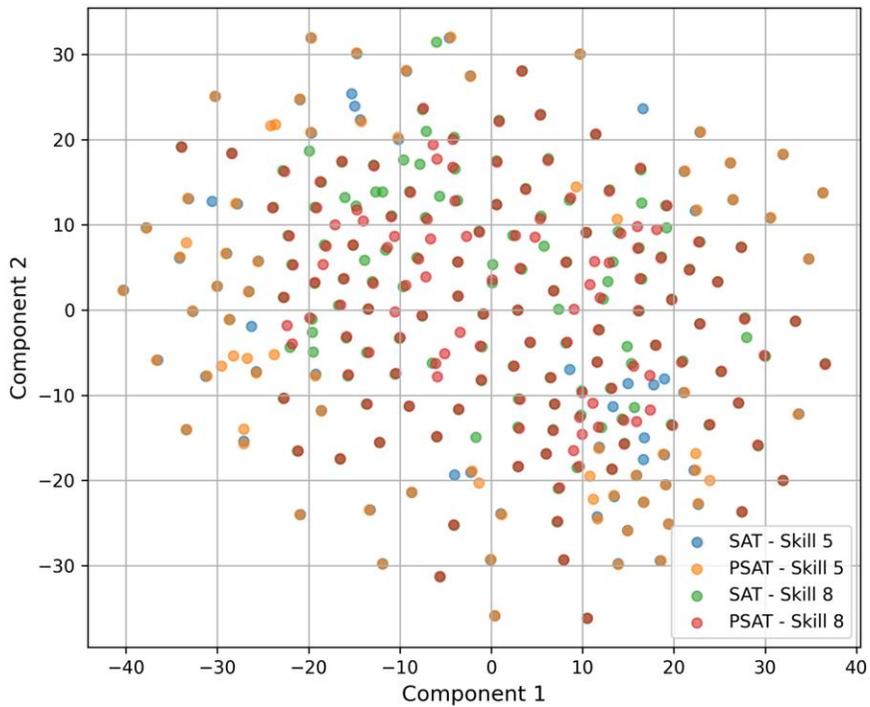

**Figure 2.** *t-SNE Plot of Embeddings for Skill 5 (Central Ideas and Details) vs. Skill 8 (Words in Context) for* Test *A and* Test *B Items*

Additionally, this study visualized item-level embeddings using dimensionality reduction techniques such as PCA, t-SNE, and ISOMAP. Figures 1 and 2 present plots using t-SNE to compare the distributions of embeddings from different skill groups for both the Test A and Test B. Plots using PCA and ISOMAP can be found in Appendix B figures. Regardless of the dimensionality reduction method used, embeddings for Skills 4 and 8, as well as Skills 5 and 8, exhibited substantial overlap. The four skill groups occupied intersecting regions in the latent space, with no distinct visual boundaries, suggesting that the items share highly similar semantic features that make it difficult for the models to correctly classify them by skill.

To further quantify these relationships, KL divergence was calculated to assess how Test B Skills 4 and 5 align distributionally with each Test A skill in the embedding space. As shown in Tables 12 and 13, Test A Skill 8 consistently exhibited low KL divergence (17.986 and 25.491), indicating high distributional similarity. These results provide additional insight into the observed misclassification patterns, in which Test B Skills 4 and 5 items were frequently predicted as Skill 8.

**Table 12**

*KL Divergence between Test B Skill 4 and Each Test A Skill*

| From | To | KL divergence |
|---|---|---|
| Test B skill 4 | Test A skill 1 | 32.927 |
| Test B skill 4 | Test A skill 2 | 38.059 |
| Test B skill 4 | Test A skill 4 | 42.588 |
| Test B skill 4 | Test A skill 4 | 44.503 |
| Test B skill 4 | Test A skill 5 | 40.996 |
| Test B skill 4 | Test A skill 6 | 13.610 |
| Test B skill 4 | Test A skill 7 | 26.869 |
| Test B skill 4 | Test A skill 8 | **17.986** |
| Test B skill 4 | Test A skill 9 | 44.342 |
| Test B skill 4 | Test A skill 10 | 74.312 |

*Note.* Skill 1 = Boundaries; Skill 2 = Form, Structure, and Sense; Skill 3 = Command of Evidence; Skill 4 = Inferences; Skill 5 = Central Ideas and Details; Skill 6 = Transitions; Skill 7 = Rhetorical Synthesis; Skill 8 = Words in Context; Skill 9 = Text Structure and Purpose; Skill 10 = Cross-Text Connections.

**Table 13**

*KL Divergence between* Test *B Skill 5 and Each* Test *A Skill*

| From | To | KL divergence |
|---|---|---|
| Test B skill 5 | Test A skill 1 | 44.096 |
| Test B skill 5 | Test A skill 2 | 48.358 |
| Test B skill 5 | Test A skill 3 | 48.800 |
| Test B skill 5 | Test A skill 4 | 65.873 |
| Test B skill 5 | Test A skill 5 | 41.134 |
| Test B skill 5 | Test A skill 6 | 44.554 |
| Test B skill 5 | Test A skill 7  | 40.371 |
| Test B skill 5 | Test A skill 8  | **25.491** |
| Test B skill 5 | Test A skill 9  | 43.649 |
| Test B skill 5 | Test A skill 10 | 83.533 |

*Note.* Skill 1 = Boundaries; Skill 2 = Form, Structure, and Sense; Skill 3 = Command of Evidence; Skill 4 = Inferences; Skill 5 = Central Ideas and Details; Skill 6 = Transitions; Skill 7 = Rhetorical Synthesis; Skill 8 = Words in Context; Skill 9 = Text Structure and Purpose; Skill 10 = Cross-Text Connections.

## 4. Conclusion and Discussion

This study demonstrated the effectiveness of fine-tuned small language models (SLMs) for automated item alignment in large-scale reading and writing assessments, using Test A and Test B items mapped to hierarchical domain and skill labels. The key findings are as follows:

First, fine-tuned SLMs substantially outperformed traditional embedding-based classifiers. Models such as ConvBERT, RoBERTa-large, and DeBERTa-base achieved near-perfect accuracy on both skill and domain alignment tasks for Test A items, while embedding-based approaches such as SVM and MLP showed lower performance, especially in fine-grained skill alignment.

Second, model performance was strongly influenced by the richness of input data. Training with more variety of input data generally outperformed the models trained with limited input data. In contrast, increasing training sample size alone yielded relatively modest gains unless paired with sufficiently rich input. Notably, the inclusion of rationales increased the token count and semantic complexity, requiring larger datasets for stable learning outcomes.

Third, generalization across assessments was partially successful. Fine-tuned models trained on Test A items transferred reasonably well to Test B items, given the shared domain-skill structure. However, several models especially BERT-base, BERT-large, RoBERTa-base, and ELECTRA-small struggled to distinguish between semantically overlapping skills such as Skill 4 (Inferences), Skill 5 (Central Ideas and Details), and Skill 8 (Words in Context). These difficulties were confirmed by cosine similarity and KL divergence analyses, which showed substantial overlap in the embedding spaces of these categories.

While the results of this study are promising, several limitations must be acknowledged. First, this study focused on single-coded items, aligned to only one domain or skill label, consistent with the current Test A Reading and Writing framework. However, real-world large-scale assessments often involve multi-coded items that align with multiple content standards. For example, Test A items have been aligned to multiple state Algebra I standards, such as those used in Arizona (Christopherson & Webb, 2020) and Florida (McCormick & Geisinger, 2017). Future research should investigate automated alignment approaches for items that are double-coded, triple-coded, or associated with even more standards, to better reflect real large-scale assessment practices.

Second, although recent studies have demonstrated the potential of prompt-based large language models (LLMs) like GPT-3.5 and GPT-4 for alignment tasks, this study did not include such models due to practical concerns particularly high computational costs and potential test security risks involved in operationalizing LLMs in educational assessment programs. In order to better understand their feasibility and effectiveness in real-world alignment tasks, future research should examine the relative strengths and limitations of prompt-based LLMs compared to fine-tuned SLMs.

Third, although the fine-tuned SLMs demonstrated high alignment accuracy, they lack interpretability. The black-box nature of these models makes it difficult to explain why a particular label was assigned to an item, which can limit stakeholder trust in high-stakes contexts. For instance, although model accuracy was high, frequent misclassification of Test B items from Skill 4 and Skill 5 as Skill 8 (see Appendix B) reveals semantic ambiguity that even advanced models fail to resolve. Future work should integrate explainability tools such as attention visualization or example-based attribution to increase transparency.

Fourth, the generalizability of SLMs across assessments remains imperfect. Although Test A-trained models generalized reasonably to Test B items, model performance dropped in skills with blurred semantic boundaries. These results underscore that even under a shared content framework, assessment-specific characteristics may affect model predictions. Partial retraining or domain adaptation strategies may be necessary to ensure cross-assessment consistency.

In conclusion, while fine-tuned SLMs represent a powerful tool for automating alignment in educational assessments, their implementation should be guided by careful attention to content structure, interpretability, model adaptability, and use-case constraints. Future research should aim to build more robust, transparent, and flexible alignment systems that are scalable to real-world educational settings.

# References

Anderson, D., Rowley, B., Stegenga, S., Irvin, P. S., & Rosenberg, J. M. (2020). Evaluating content-related validity evidence using a text-based machine learning procedure. *Educational Measurement: Issues and Practice*, *39*(4), 53-64.

Bhola, D. S., Impara, J. C., & Buckendahl, C. W. (2003). Aligning tests with states' content standards: Methods and issues. *Educational Measurement: Issues and Practice, 22*(3), 21–29.

Bier, N., Moore, S., & Van Velsen, M. (2019, March). Instrumenting courseware and leveraging data with the Open Learning Initiative (OLI). In *Proceedings of the 9th International Conference on Learning Analytics & Knowledge (LAK19)*

Butterfuss, R., & Doran, H. (2025). An application of text embeddings to support alignment of educational content standards. *Educational Measurement: Issues and Practice, 44*(1), 73–83.

Camilli, G. (2024). An NLP crosswalk between the Common Core State Standards and NAEP item specifications. *arXiv preprint arXiv:2405.17284*.

Christopherson, S. C., & Webb, N. L. (2020). *Alignment Analysis of Two Forms of the SAT with the Arizona Academic Standards for English Language Arts Grades 11–12, Algebra 1, and Geometry*. Wisconsin Center for Education Products and Services.

Clark, K., Luong, M. T., Le, Q. V., & Manning, C. D. (2020). Electra: Pre-training text encoders as discriminators rather than generators. *arXiv preprint arXiv:2003.10555*.

Cui, Y., Che, W., Liu, T., Qin, B., Wang, S., & Hu, G. (2020). Revisiting pre-trained models for Chinese natural language processing. In T. Cohn, Y. He, & Y. Liu (Eds.), *Findings of the Association for Computational Linguistics: EMNLP 2020* (pp. 657–668). Association for Computational Linguistics. https://doi.org/10.18653/v1/2020.findings-emnlp.58

Devlin, J., Chang, M. W., Lee, K., & Toutanova, K. (2019). BERT: Pre-training of deep bidirectional transformers for language understanding. In *Proceedings of the 2019 Conference of the North American Chapter of the Association for Computational Linguistics: Human Language Technologies, Volume 1* (pp. 4171–4186).

Ding, Z., Wang, X., Wu, Y., Cao, G., & Chen, L. (2025). Tagging knowledge concepts for math problems based on multi-label text classification. *Expert Systems with Applications, 267*, 126232.

Dodge, J., Ilharco, G., Schwartz, R., Farhadi, A., Hajishirzi, H., & Smith, N. (2020). Fine-tuning pretrained language models: Weight initializations, data orders, and early stopping. *arXiv preprint arXiv:2002.06305*.

Embretson, S. E., & Reise, S. P. (2000). *Item response theory for psychologists*. Psychology Press.

He, P., Gao, J., & Chen, W. (2021). Debertav3: Improving deberta using electra-style pre-training with gradient-disentangled embedding sharing. *arXiv preprint arXiv:2111.09543*.

He, P., Liu, X., Gao, J., & Chen, W. (2020). Deberta: Decoding-enhanced bert with disentangled attention. *arXiv preprint arXiv:2006.03654*.

Herman, J. L., Webb, N. M., & Zuniga, S. (2003). *Alignment and college admissions: The match of expectations, assessments, and educator perspectives*. Center for the Study of Evaluation, CRESST, UCLA.

Huang, T., Hu, S., Yang, H., Geng, J., Liu, S., Zhang, H., & Yang, Z. (2023). PQSCT: Pseudo-Siamese BERT for concept tagging with both questions and solutions. *IEEE Transactions on Learning Technologies, 16*(5), 831–846. https://doi.org/10.1109/TLT.2023.3275707

Jiang, Z. H., Yu, W., Zhou, D., Chen, Y., Feng, J., & Yan, S. (2020). Convbert: Improving bert with span-based dynamic convolution. *Advances in Neural Information Processing Systems*, *33*, 12837-12848.

Kane, M. (2006). Content-Related Validity Evidence in Test Development. In S. M. Downing & T. M. Haladyna (Eds.), *Handbook of test development* (pp. 131–153). Lawrence Erlbaum Associates.

Karlovčec, M., Córdova-Sánchez, M., & Pardos, Z. A. (2012). Knowledge component suggestion for untagged content in an intelligent tutoring system. In S. A. Cerri, W. J. Clancey, G. Papadourakis, & K. Panourgia (Eds.), *Intelligent tutoring systems: 11th International Conference, ITS 2012, Chania, Crete, Greece, June 14–18, 2012. Proceedings* (Lecture Notes in Computer Science, Vol. 7315, pp. 195–200). Springer. https://doi.org/10.1007/978-3-642-30950-2_25

Khan, S., Rosaler, J., Hamer, J., & Almeida, T. (2021). Catalog: An educational content tagging system. In *Proceedings of the 14th International Conference on Educational Data Mining (EDM 2021)*. International Educational Data Mining Society.

Kim, Y. (2014). Convolutional neural networks for sentence classification. In A. Moschitti, B. Pang, & W. Daelemans (Eds.), In *Proceedings of the 2014 Conference on Empirical Methods in Natural Language Processing (EMNLP)* (pp. 1746–1751). Association for Computational Linguistics. https://doi.org/10.3115/v1/D14-1181

Lan, Z., Chen, M., Goodman, S., Gimpel, K., Sharma, P., & Soricut, R. (2019). Albert: A lite bert for self-supervised learning of language representations. *arXiv preprint arXiv:1909.11942*.

Li, H., Xu, T., Tang, J., & Wen, Q. (2024). Automate knowledge concept tagging on math questions with LLMs. *arXiv preprint arXiv:2403.17281*.

Lima, P. S. N., Ambrosio, A. P., Felix, I., Brancher, J. D., & de Carvalho, D. T. (2018). Content analysis of student assessment exams. In *2018 IEEE Frontiers in Education Conference (FIE)* (pp. 1–9). IEEE. https://doi.org/10.1109/FIE.2018.8659169

Liu, N., Sonkar, S., Basu Mallick, D., Baraniuk, R., & Chen, Z. (2025). Atomic learning objectives and LLMs labeling: A high-resolution approach for physics education. In *Proceedings of the 15th International Learning Analytics and Knowledge Conference (LAK '25)* (pp. 620–630). Association for Computing Machinery. https://doi.org/10.1145/3706468.3706550

Liu, Y., Ott, M., Goyal, N., Du, J., Joshi, M., Chen, D., ... & Stoyanov, V. (2019). Roberta: A robustly optimized bert pretraining approach. *arXiv preprint arXiv:1907.11692*.

Martone, A., & Sireci, S. G. (2009). Evaluating alignment between curriculum, assessment, and instruction. *Review of Educational Research, 79*(4), 1332–1361.

McCormick, C., & Geisinger, K. F. (2017). *Alignment Study Full Report*. Buros Center for Testing.

Mikolov, T., Chen, K., Corrado, G., & Dean, J. (2013). Efficient estimation of word representations in vector space. *arXiv preprint arXiv:1301.3781*.

Moore, S., Schmucker, R., Mitchell, T., & Stamper, J. (2024). *Automated generation and tagging of knowledge components from multiple-choice questions*. In *Proceedings of the Eleventh ACM Conference on Learning @ Scale (L@S '24)* (pp. 122–133). Association for Computing Machinery. https://doi.org/10.1145/3657604.3662030

Mosbach, M., Andriushchenko, M., & Klakow, D. (2020). On the stability of fine-tuning bert: Misconceptions, explanations, and strong baselines. *arXiv preprint arXiv:2006.04884*.

Muennighoff, N., Tazi, N., Magne, L., & Reimers, N. (2022). MTEB: Massive text embedding benchmark. *arXiv preprint arXiv:2210.07316*.

Nemeth, Y., Michaels, H., Wiley, C., & Chen, J. (2016). *Delaware System of Student Assessment and Maine Comprehensive Assessment System: SAT alignment to the Common Core State Standards – Final Report*. Human Resources Research Organization.

Ozyurt, Y., Feuerriegel, S., & Sachan, M. (2025). Automated knowledge concept annotation and question representation learning for knowledge tracing. *arXiv Preprint*, *arXiv:*2410.01727. https://doi.org/10.48550/arXiv.2410.01727

Pardos, Z. A., & Dadu, A. (2017). Imputing KCs with representations of problem content and context. *In Proceedings of the 25th Conference on User Modeling, Adaptation and Personalization (UMAP '17)* (pp. 148–155). Association for Computing Machinery. https://doi.org/10.1145/3079628.3079689

Pennington, J., Socher, R., & Manning, C. D. (2014). GloVe: Global vectors for word representation. In *Proceedings of the 2014 Conference on Empirical Methods in Natural Language Processing (EMNLP)* (pp. 1532–1543).

Peters, S., Zhang, N., Jiao, H., Li, M., Zhou, T., Lissitz, R., Fu, Y., & Xu, Q. (2025). *Text-based approaches to item difficulty modeling in high-stakes assessments: A systematic review* (MARC Research Report). University of Maryland.

Qu, B., Cong, G., Li, C., Sun, A., & Chen, H. (2012). An evaluation of classification models for question topic categorization. *Journal of the American Society for Information Science and Technology*, *63*(5), 889-903.

Ramesh, R., Sasikumar, M., & Iyer, S. (2016). A software tool to measure the alignment of assessment instrument with a set of learning objectives of a course. In *2016 IEEE 16th International Conference on Advanced Learning Technologies (ICALT)* (pp. 64–68). IEEE. https://doi.org/10.1109/ICALT.2016.10

Reimers, N., & Gurevych, I. (2021). *all-distilroberta-v1* [Computer software]. Hugging Face. https://huggingface.co/sentence-transformers/all-distilroberta-v1

Sanh, V., Debut, L., Chaumond, J., & Wolf, T. (2019). DistilBERT, a distilled version of BERT: smaller, faster, cheaper and lighter. *arXiv preprint arXiv:1910.01108*.

Schuster, M., & Paliwal, K. K. (1997). Bidirectional recurrent neural networks. *IEEE Transactions on Signal Processing, 45*(11), 2673–2681.

Sebastiani, F. (2002). Machine learning in automated text categorization. *ACM computing surveys (CSUR)*, *34*(1), 1-47.

Shen, J. T., Yamashita, M., Prihar, E., Heffernan, N., Wu, X., McGrew, S., & Lee, D. (2021). Classifying math knowledge components via task-adaptive pre-trained BERT. In *Artificial Intelligence in Education: 22nd International Conference, AIED 2021, Utrecht, The Netherlands, June 14–18, 2021, Proceedings, Part I 22* (pp. 408-419). Springer International Publishing.

Smith, M. S., & O'Day, J. (1990). Systemic school reform. *Journal of Education Policy, 5*(5), 233–267. https://doi.org/10.1080/02680939008549074

Sparck Jones, K. (1972). A statistical interpretation of term specificity and its application in retrieval. *Journal of documentation*, *28*(1), 11-21.

Sun, B., Zhu, Y., Xiao, Y., Xiao, R., & Wei, Y. (2018). Automatic question tagging with deep neural networks. *IEEE Transactions on Learning Technologies*, *12*(1), 29-43.

Tan, C. S., & Kim, J. J. (2024). Automated Math Word Problem Knowledge Component Labeling and Recommendation. In *International Conference in Methodologies and *intelligent Systems for Technology Enhanced Learning* (pp. 338-348). Cham: Springer Nature Switzerland.

Tian, Z., Flanagan, B., Dai, Y., & Ogata, H. (2022). Automated matching of exercises with knowledge components. In *30th International Conference on Computers in Education Conference Proceedings* (pp. 24-32).

Wang, L., Yang, N., Huang, X., Yang, L., Majumder, R., & Wei, F. (2024). Multilingual e5 text embeddings: A technical report. *arXiv preprint arXiv:2402.05672*.

Wang, T., Stelter, K., Floyd, J., O'Neill, T., Hendrix, N., Bazemore, A., Rode, K., & Newton, W. (2023). *Blueprinting the future: Automatic item categorization using hierarchical zero-shot and few-shot classifiers*. arXiv. https://arxiv.org/abs/2312.03561

Webb, N. L. (1997). *Criteria for alignment of expectations and assessments in mathematics and science education*. Research Monograph No. 6.

Yilmazel, O., Balasubramanian, N., Harwell, S. C., Bailey, J., Diekema, A. R., & Liddy, E. D. (2007). Text categorization for aligning educational standards. *In 2007 40th Annual Hawaii International Conference on System Sciences (HICSS'07)* (p. 73). IEEE. https://doi.org/10.1109/HICSS.2007.517

Yu, R., Das, S., Gurajada, S., Varshney, K., Raghavan, H., & Lastra-Anadon, C. (2021). A research framework for understanding education-occupation alignment with NLP techniques. In A. Field, S. Prabhumoye, M. Sap, Z. Jin, J. Zhao, & C. Brockett (Eds.), *Proceedings of the 1st Workshop on NLP for Positive Impact* (pp. 100–106). Association for Computational Linguistics. https://doi.org/10.18653/v1/2021.nlp4posimpact-1.11

Zhang, N., Jiao, H., Yadav, C., & Lissitz, R. (2025). *Aligning SAT math to state math content standards: A systematic review* (Technical report). Maryland Assessment Research Center, University of Maryland.

Zhou, Z., & Ostrow, K. S. (2022). Transformer-based automated content-standards alignment: A pilot study. In G. Meiselwitz (Ed.), *HCI International 2022 – Late Breaking Papers: Interaction in New Media, Learning and Games* (Vol. 13517, pp. 525–542). Springer. https://doi.org/10.1007/978-3-031-22131-6_39

# Appendix A

**Table A.1**

*Content Domains*

| Domain | Description |
|---|---|
| Standard English Conventions | Assesses the ability to identify and correct errors related to grammar, punctuation, and sentence structure in written English. |
| Information and Ideas | Evaluates the ability to extract, interpret, and integrate information from texts and visual data (e.g., graphs or tables), and to make analytical judgments based on that information. |
| Expression of Ideas | Measures students' ability to improve the clarity, coherence, and rhetorical effectiveness of written texts by revising them to achieve specific communicative goals. |
| Craft and Structure | Assesses students' ability to interpret context-dependent vocabulary, analyze rhetorical strategies, and draw meaningful connections between thematically related texts. This domain emphasizes a deeper understanding of how language and text structure contribute to an author's purpose or argument. |

**Table A.2**

*Question Text Templates by Domain and Skill.*

| Domain | Skill | Question Text |
|---|---|---|
| Standard English Conventions | Boundaries | Which choice completes the text so that it conforms to the conventions of Standard English? |
| | Form, Structure, and Sense | |
| Information and Ideas | Command of Evidence | Which choice best describes data in/from table/graph that support …? <br> Which choice most effectively uses data from the table/graph to complete …? <br> Which finding about/from the …, if true, would most directly support? <br> Which finding, if true, would most directly support/weaken …? <br> Which quotation from … most effectively illustrates the claim? <br> Which statement, if true, would most strongly support the claim …. |
| | Inferences | Which choice most logically completes the text? |
| | Central Ideas and Details | According to the text, ... <br> Based on the text, … <br> What does the text indicate/most strongly suggest …? <br> Which choice best states the main idea of the text? |
| Expression of Ideas | Transitions | Which choice completes the text with the most logical transition? |

| | Rhetorical Synthesis | The student wants to …, Which choice most effectively uses relevant information from the notes to accomplish this goal? |
|---|---|---|
| Craft and Structure | Words in Context | As used in the text, what does the word …most nearly mean? |
| | Text Structure and Purpose | Which choice best describes/states the main purpose/function of the … |
| | Cross-Text Connections | Based on the texts, how would … Which choice best describes a difference in how the author of Text 1 and the author of Text 2 view … |

**Table A.3**
*Example of a Preprocessed Training Sample.*

| Question ID | Inputs | Domain | Skill |
|---|---|---|---|
| 267a13e2 | In 2010, archaeologist Noel Hidalgo Tan was visiting the twelfth-century temple of Angkor Wat in Cambodia when he noticed markings of red paint on the temple ______ the help of digital imaging techniques, he discovered the markings to be part of an elaborate mural containing over 200 paintings. A: walls, with \| B: walls with \| C: walls so with \| D: walls. With D Choice D is the best answer. The convention being tested is punctuation use between sentences. In this choice, the period after "walls" is used correctly to mark the boundary between the first sentence ("In...walls") and the second sentence ("With…techniques"), which starts with a supplementary phrase. Choice A is incorrect because it results in a comma splice. A comma can't be used in this way to mark the boundary between sentences. Choice B is incorrect because it results in a run-on sentence. The sentences ("In...walls" and "with...paintings") are fused without punctuation and/or a conjunction. Choice C is incorrect. Without a comma preceding it, the conjunction "so" can't be used in this way to join sentences. | Standard English Conventions | Boundaries |

*Note.* Yellow indicates the prompt, green indicates the answer options, blue marks the correct answer/key, and bisque color highlights the rationale.

**Appendix B**

To better identify where misclassifications occurred, Figures B.1 and B.2 present the confusion matrices for skill alignment, while Figures B.3 and B.4 display the confusion matrices for domain alignment. The BERT-base model was used as a reference for comparison against the best-performing models in both skill and domain alignment tasks. Figure B.1 shows that the BERT-base model frequently misclassified Test B items from Skill 4 (*Inferences*) and Skill 5 (*Central Ideas and Details*) as Skill 8 (*Words in Context*). In contrast, the best-performing RoBERTa-large model, as depicted in Figure B.2, misclassified only 6 out of 1,052 Test B items.

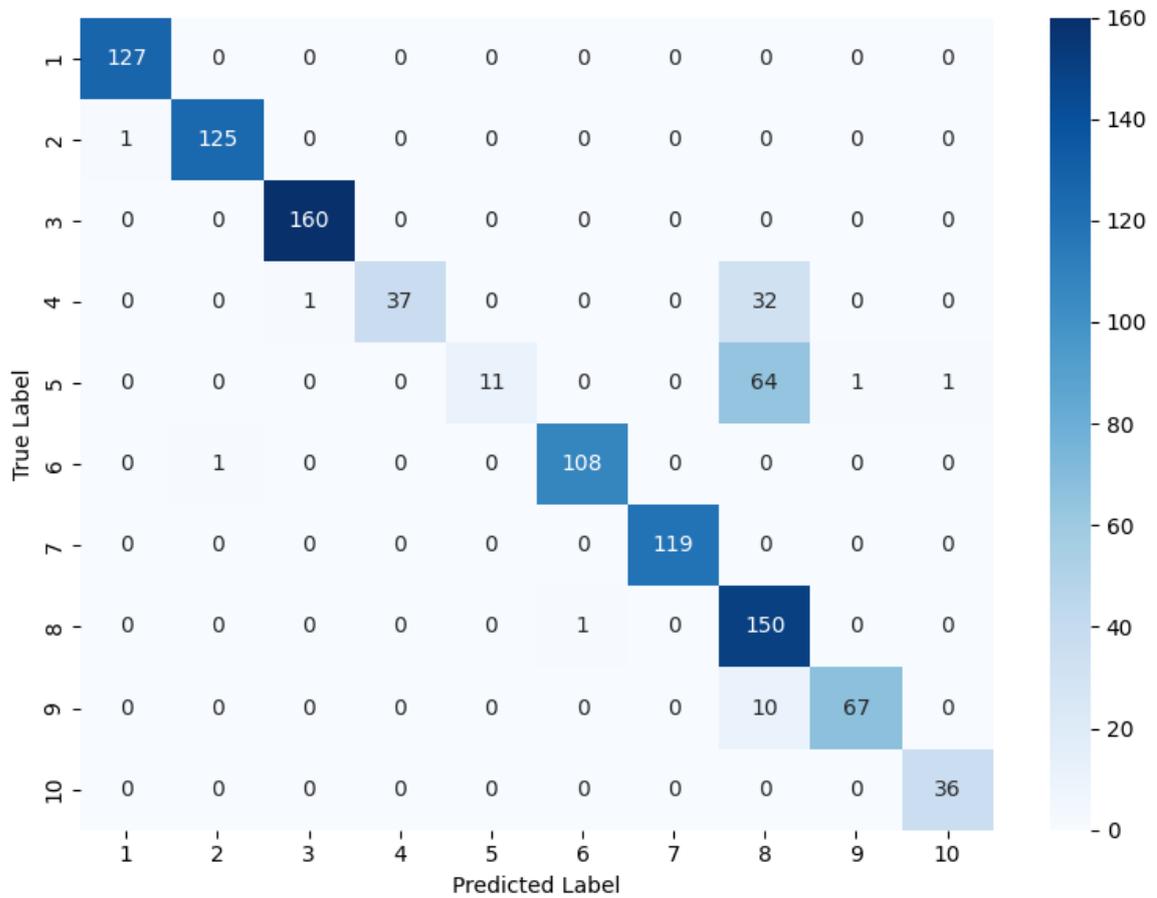

**Figure B.1.** *Confusion Matrix of the BERT-base Model for* Test *B Skill Alignment*

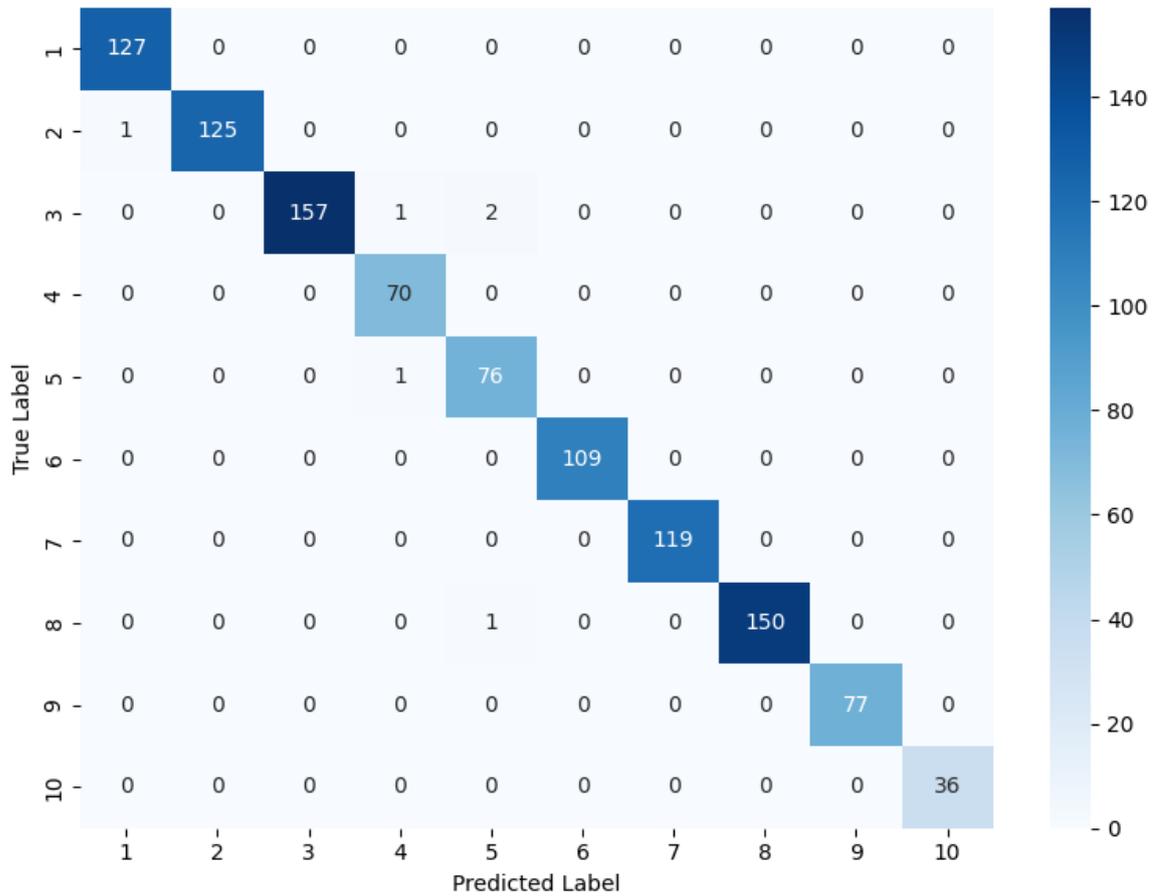

**Figure B.2.** *Confusion Matrix of the RoBERTa-large Model for* Test *B Skill Alignment*

The confusion matrices for domain alignment are presented in Figures B.3 and B.4. As shown in Figure B.3, the BERT-base model frequently misclassified Test B items from Domain 2 (*Information and Ideas*) as Domain 4 (*Craft and Structure*). In contrast, the best-performing DeBERTa-base model misclassified only 3 out of 1,052 Test B items, as illustrated in FigureB.4.

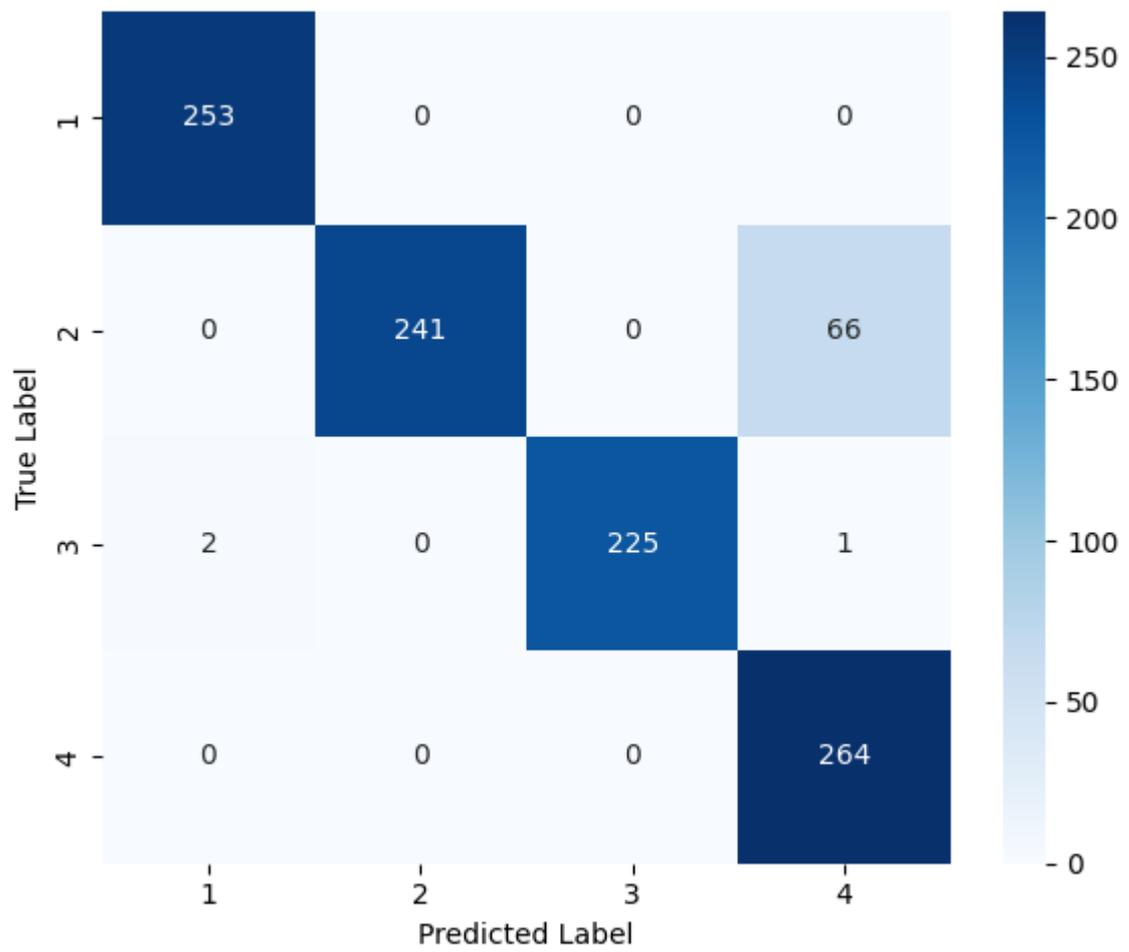

**Figure B.3.** *Confusion Matrix of the BERT-base Model for* Test *B Domain Alignment*

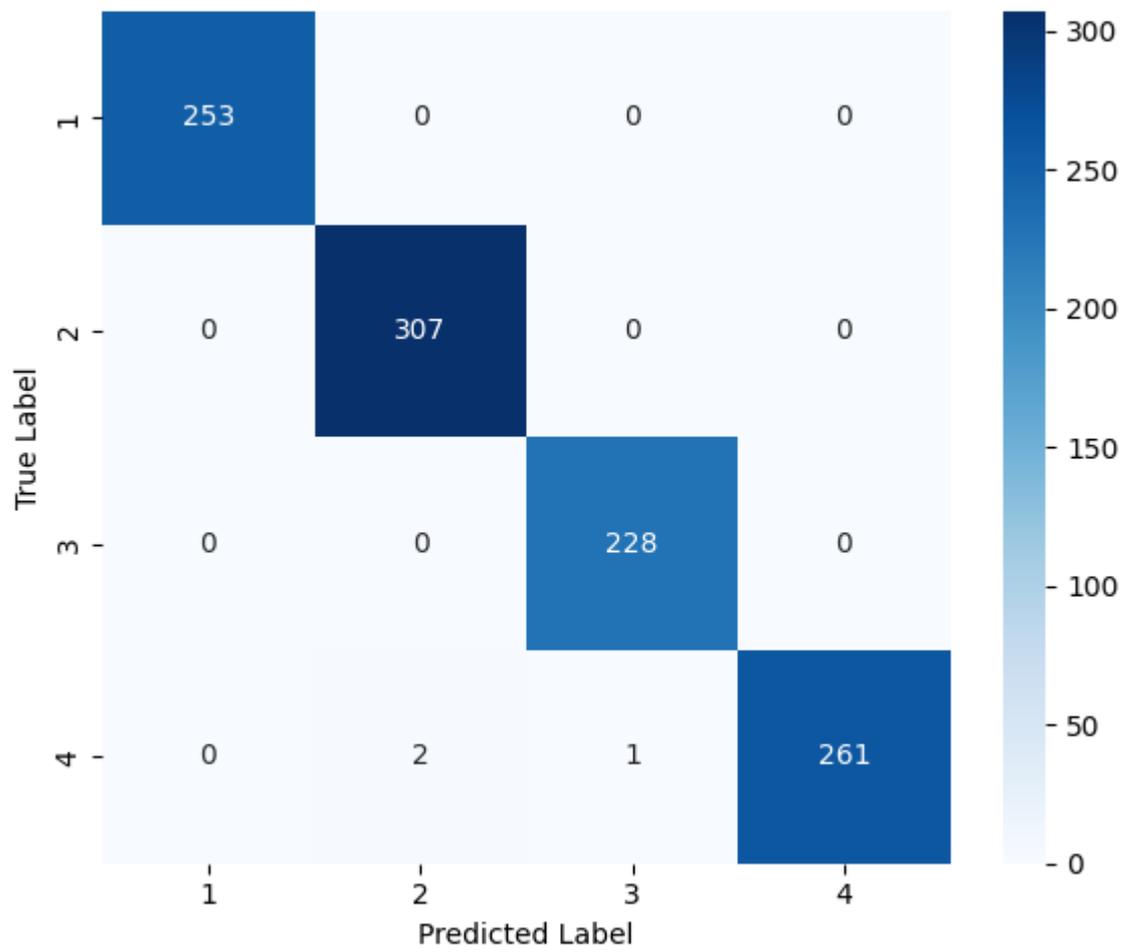

**Figure B.4.** *Confusion Matrix of the DeBERTa-base Model for* Test *B Domain Alignment Task*

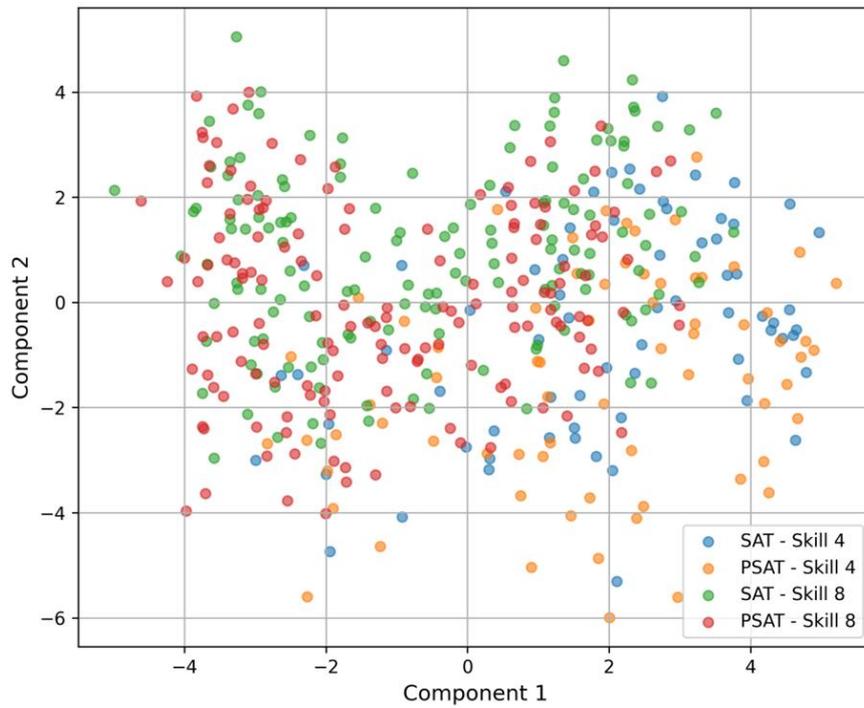

**Figure B.5.** *PCA Plot of Embeddings for Skill 4 (Inferences) vs. Skill 8 (Words in Context) for* Test *A and* Test *B Items*

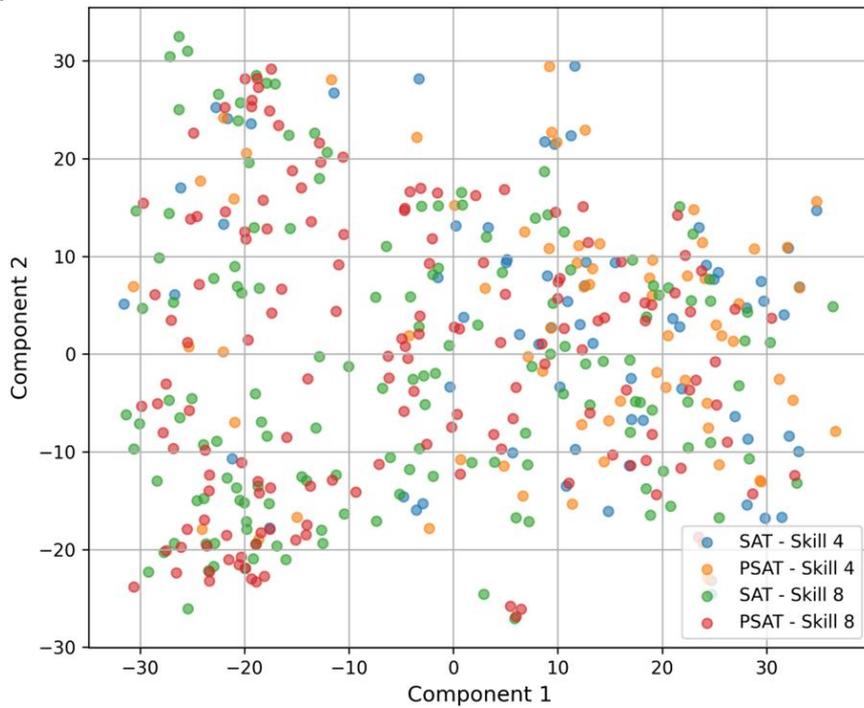

**Figure B.6.** *ISOMAP Plot of Embeddings for Skill 4 (Inferences) vs. Skill 8 (Words in Context) for* Test *A and* Test *B Items*

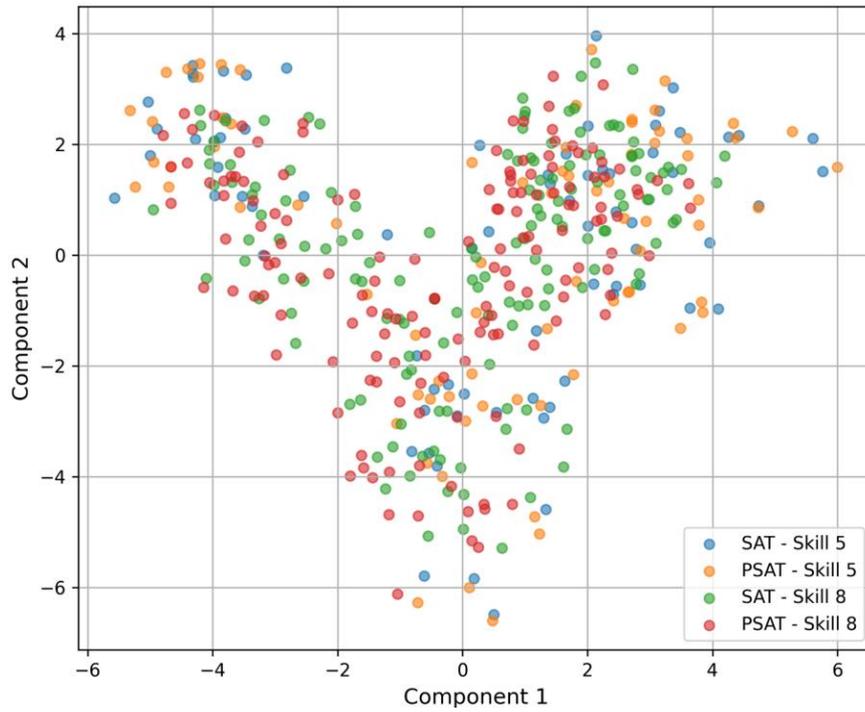

**Figure B.7.** *PCA Plot of Embeddings for Skill 5 (Central Ideas and Details) vs. Skill 8(Words in Context) for* Test *A and* Test *B Items*

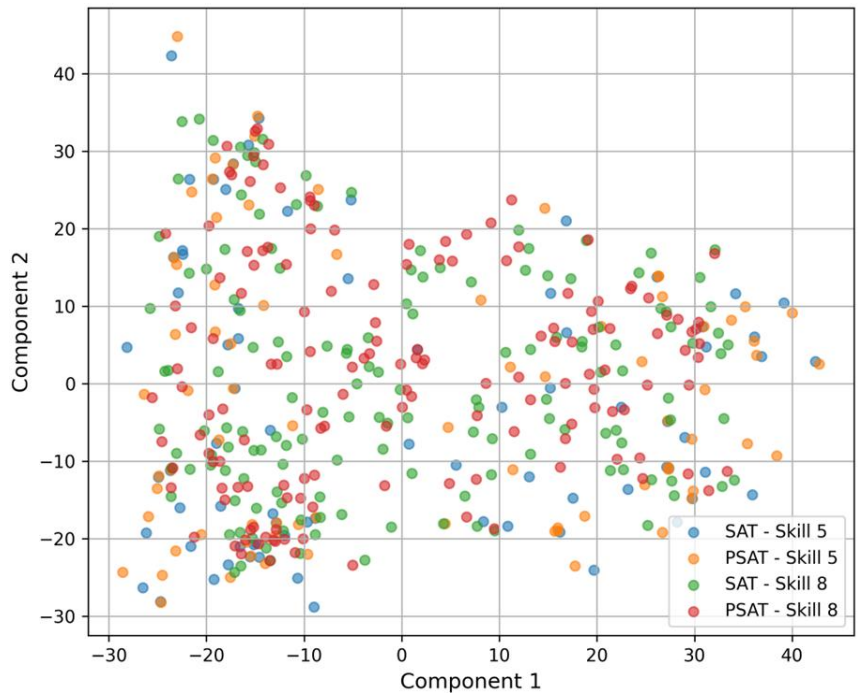

**Figure B.8.** *ISOMAP Plot of Embeddings for Skill 5 (Central Ideas and Details) vs. Skill 8 (Words in Context) for* Test *A and* Test *B Items*